\documentclass[10pt,twocolumn]{article}

\usepackage[utf8]{inputenc}

\usepackage{graphicx}
\usepackage{charter}
\usepackage[hyphens]{url}
\usepackage{hyperref}
\usepackage[english]{babel}

\usepackage{natbib}
\setcitestyle{authoryear,open={(},close={)}} 

\title{Who could be behind QAnon? \\%
Authorship attribution with supervised machine-learning%
}
\date{}

\usepackage{authblk}

\author[1]{Florian Cafiero (\textsc{orcid} 0000-0002-1951-6942)}
\author[2]{Jean-Baptiste Camps (\textsc{orcid} 0000-0003-0385-7037)}

\affil[1]{Sciences Po, Medialab, 1 place Saint Thomas d'Aquin, Paris, 75007, France}
\affil[2]{École nationale des chartes, Université Paris, Sciences \& Lettres, 65 rue de Richelieu, Paris, 75002, France}

\makeatletter
\renewcommand{\paragraph}{%
  \@startsection{paragraph}{4}%
  {\z@}{1.25ex \@plus 1ex \@minus .2ex}{-1em}%
  {\normalfont\normalsize\bfseries}%
}
\makeatother

\begin{document}

\maketitle

\begin{abstract}
A series of social media posts signed under the pseudonym ``Q'', started a movement known as QAnon, which led some of its most radical supporters to violent and illegal actions. To identify the person(s) behind Q, we evaluate the coincidence between the linguistic properties of the texts written by Q and to those written by a list of suspects provided by journalistic investigation. To identify the authors of these posts, serious challenges have to be addressed. The “Q drops” are very short texts, written in a way that constitute a sort of literary genre in itself, with very peculiar features of style. These texts might have been written by different authors, whose other writings are often hard to find. After an online ethnology of the movement, necessary to collect enough material written by these thirteen potential authors, we use supervised machine learning to build stylistic profiles for each of them. We then performed a rolling analysis on Q's writings, to see if any of those linguistic profiles match the so-called `QDrops' in part or entirety. We conclude that two different individuals, Paul F. and Ron W., are the closest match to Q's linguistic signature, and they could have successively written Q's texts. These potential authors are not high-ranked personality from the U.S. administration, but rather social media activists.
\end{abstract}

\section*{Introduction}

The QAnon movement revolves around the posts of one or more individuals signing their message under the name ``Q'' on online forums. These messages first appeared in October 2017 on 4chan, a popular imageboard known for its appreciation of the anime culture, its memes, but also for its \textbackslash pol\textbackslash board, dedicated to ``politically incorrect'' contents. Q's messages were later on posted on 8chan (now 8kun), a competing imageboard created after 4chan, considered as a place allowing even more ``free-speech'', and in particular, letting users easily create a ``board'' regarding the topic of their choice \citep{baele2021variations}. But who wrote these messages? The QAnon supporters see in Q one or a few top-rank personalities around Donald Trump, maybe Trump himself. Journalistic investigations point on the contrary to social media activists \citep{zadrozny_who_2018}.

\section*{Why work on QAnon? Specificities and social impact}

The violence of the QAnon believers is not unprecedented nor unmatched
on social networks. For instance, discussions around QAnon on Voat, a Reddit-like forum attracting many alt-right and QAnon supporters, even show less toxicity than elsewhere in average on the rest of this admittedly very specific platform \citep{papasavva2021qoincidence}. The singularity of this online group mostly resides in how much
its theories have spread, and on its important consequences in real
life. Considered a potential domestic terrorist threat by the FBI since 2018 \citep{winter2019exclusive},
the most radical QAnon supporters were implicated in a variety of
criminal incidents and violent events
\citep{garry2021qanon}.
Recent research has shown that, beyond political orientation, the main
common trait between people arrested for breaking into the United States
Capitol in Washington on January 6, 2021
\citep{kaplan2021conspiracy} was
that they believed in the QAnon narrative. Documented in the press
\citep{gilbert2021qanon}, the
high impact of the QAnon theories on the social life of its believers has been compared to the ``conversion to a cult or
high-pressure religious group''
\citep{kaplan2021conspiracy}. QAnon
also affected people far beyond the U.S. borders, spreading in
particular through social media and instant messaging software. A
large-scale study on QAnon on Telegram for instance showed that messages
in German even outnumbered at some point the posts written in English,
and that posts in German and Portuguese used an even more ``toxic''
language \citep{hoseini2021globalization}.

\section*{Who is Q? The theories put to test}

Theories regarding the author behind the posts signed ``Q'' fit into two
main categories. Believers in the authenticity of the source argue that
a single source or a collective of people from U.S. intelligence
agencies would have authored the various posts. Persons cited as a
plausible author or co-author of Q range from General Michael Flynn
to Donald J. Trump or his
entourage \citep{QHBO}.
A second group of theory states that one person would have posted as Q, but without having any specific access to exclusive and reliable
sources. NBC reporters claimed that they had traced the success of QAnon to three people: `Pamphlet Anon', a.k.a. Coleman R., `BaruchtheScribe', a.k.a. Paul F., and `Tracy Beanz', a.k.a. Tracy D.\citep{zadrozny_who_2018}. Frederick Brennan, inventor of 8chan, claims that Ron and John W. are paying someone to carry on as Q, or are even acting as Q themselves \citep{QHBO}. The third group of theories holds that Q is a collective, with a small number of people sharing access to the account. This third category ``includes the notion that Q is a new kind of open-source military-intelligence agency'' \citep{lafrance_prophecies_2020}.

\section*{Authorship attribution}

To help understanding who might hide behind the pseudonym Q, we rely here on authorship attribution techniques, also known as \textit{stylometry} \citep{juola2008authorship,cafiero2022affaires}, a term coined at the turn of the 19th century to designate the measure of stylistic affinity between texts \citep{lutoslawski_principes_1898}. 
Stylometry tries to spot the idiosyncratic properties of someone's language, called \textit{idiolect} \citep{coulthard2004author}. To identify a linguistic signature of a person, regardless of the topic of the text it produces, stylometry focuses on deep linguistic properties, less subject to conscious manipulation or to context variation. Linguistic feature such as functions words (``and'', ``or'', ``upon'' etc.) \citep{kestemont2014function, segarra2015authorship}, punctuation \citep{jin2012text},  patterns of ``parts-of-speech'' (e.g. `Proper noun / Verb / Article', `Article / common noun / verb') \citep{bjorklund2017syntactic} are for instance used in specific combinations by individuals, and are a reliable clue of who is speaking or writing. 
Studies in psycholinguistics have since shown that functions words and grammatical markers are processed differently by the brain than lexical content words, and are indeed not only revealing of individual use, but also correlate to socio-cultural categories such as gender, age group or native language for instance \citep{argamon_automatically_2009,pennebaker_secret_2013}.
Since a famous study on the authors of the \textit{Federalist papers} \citep{mosteller1963inference}, a review advocating the adoption of the future U.S. constitution, many applications have been made to solve literary or historical controversies, such as Caesar's contribution to the \textit{Commentaries} on his wars \citep{kestemont2016authenticating}, or Shakespeare \citep{plechavc2019relative} and Molière's \citep{cafiero2019moliere, cafiero2021psyche} disputed authorships. It was also used to reveal who may have written Elena Ferrante's novels \citep{eder2018elena, rybicki2018partners, mikros2017blended} or to recognize J.K. Rowling's style behind the pseudonym Robert Galbraith \citep{juola2015rowling}. Applications to forensic cases have grown in the past decades, being and being more and more used in U.S. courts \citep{chaski2005s} among other, and applied to a wide range of cases ranging from immigration disputes \citep{juola2012stylometry} to murder investigations \citep{cafiero2022affaires}.

Previous research in this field suggested that there could be more than one author to the Q drops. Stylometric analysis, based on factor analysis on character 3-grams, suggested that there were probably two authors who wrote these texts, one after the other \citep{roten_style_2020}. Analyses of the distribution of the number of character and words in each Qdrops, as well as the use of special characters suggested that two hands could have written the Q drops \citep{aliapoulios2021gospel}, posts written under one of the 10 tripcodes Q used exhibiting different properties from the rest of them. Examinations of the pictures posted by Q however show that they have mostly been posted from the same Time Zone in Asia, and that original pictures were taken from the same camera all along the period \citep{Pictures}. This is interpreted as a sign that they were would one  unique author to the Q drops .

\section*{Results}

\begin{figure*}[!htb]
    \centering
    \includegraphics[width=11.4cm]{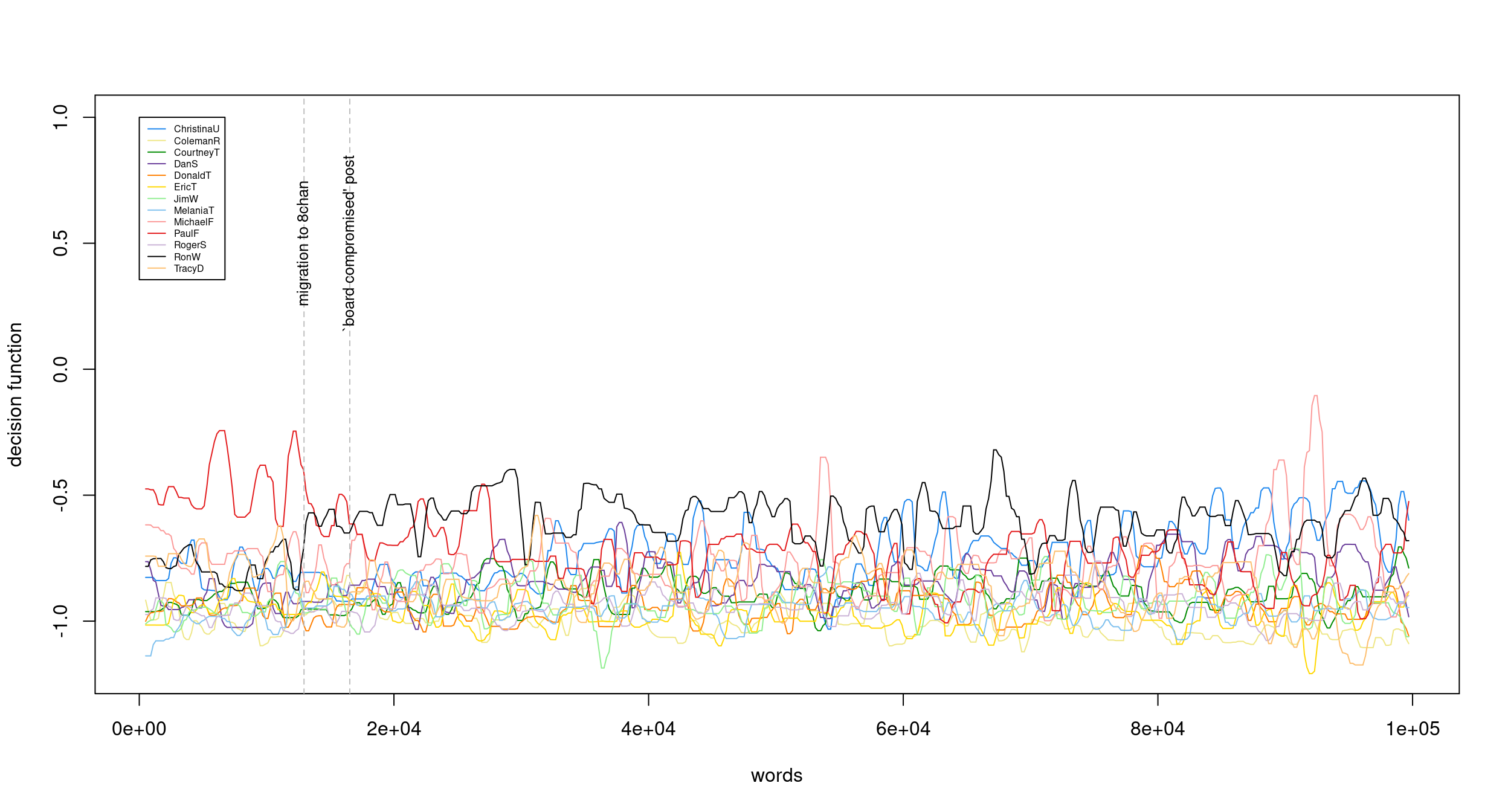}
    \includegraphics[width=11.4cm]{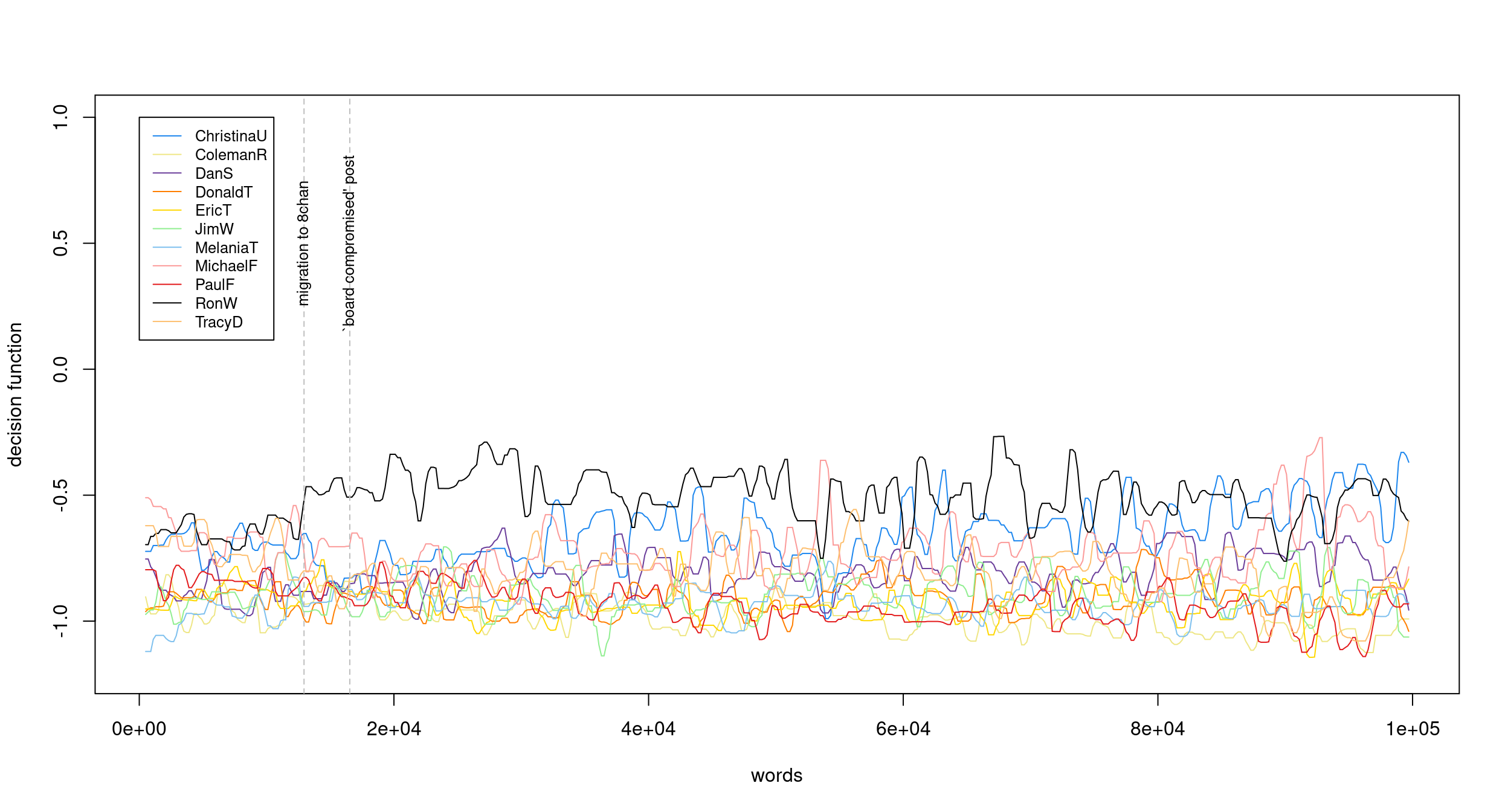}
    \caption{Decision function of each
classifier for each successive overlapping windows of Q drops (windows of 1000 words, with steps of 200 words), arranged
in chronological order for the large corpus (top) and the controlled corpus (bottom)}
    \label{fig:results}
\end{figure*}

Profiles capturing unconscious features of style such as grammatical morphemes have been built from two corpus of texts (a large corpus with all 13 candidates, but sometimes a low amount of relevant material for some of them, and a smaller, more controlled corpus) signed by each putative authors,
using supervised machine learning, with a general attributive performance of over 97\% (\textit{Materials and methods}). 
They 
show that, for most of the slices, the highest decision function is by far by Ron W.  
 (Fig.~\ref{fig:results}).
The most significant deviation from this concerns the first period of the QDrops, before the switch to 8chan. In this period, the larger corpus analysis gives Paul F. as, by far, the top candidate, before a period where Paul F. and Ron W. signals are competing, until finally Ron W. signals takes over, after a second break that closely matches a tweet described by Paul F. himself as the last authentic Qdrop, that goes
\begin{quotation} \itshape
\noindent There will be no further posts on this board under this ID.\\This will verify the trip is safeguarded and in our control.\\ This will verify this board is compromised.\\ God bless each and every one of you. \\ Fight, fight, fight!\\
Q
\end{quotation}
The dominance of Paul F. in the first period is not seen at all on the smaller corpus analysis. 

More secondarily, there are very localised spikes of Christina U. and Michael F. signals, especially in the more recent period of the QDrops.  
The rest of the candidates lag far behind.

\begin{figure*}[!htb]
    \centering
    \includegraphics[width=0.48\textwidth]{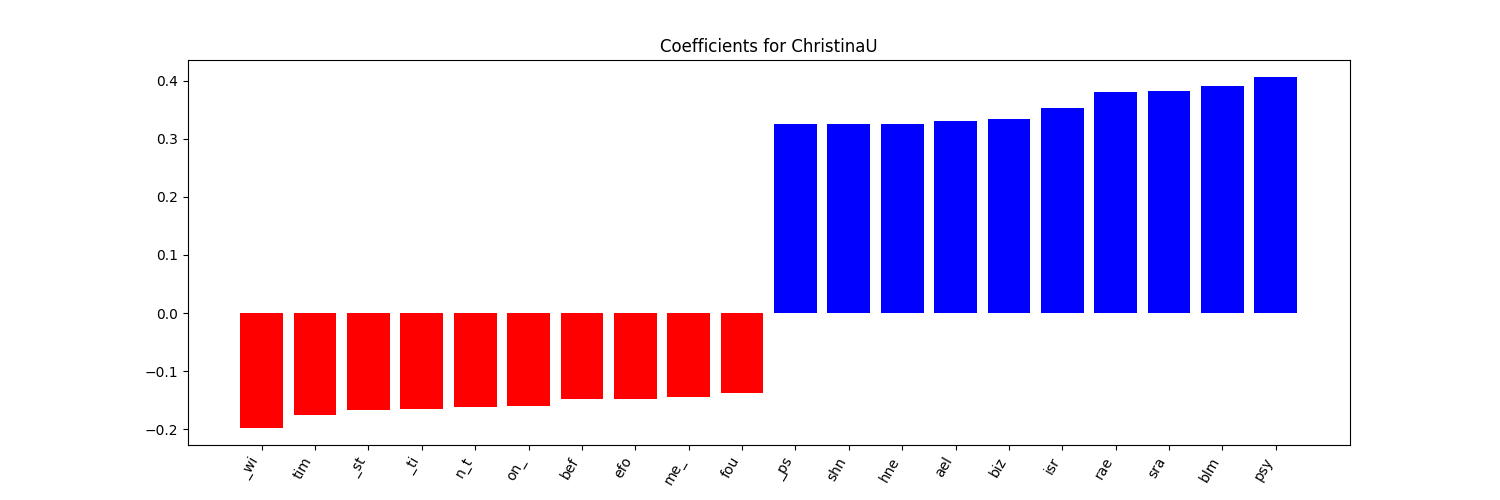}
    \includegraphics[width=0.48\textwidth]{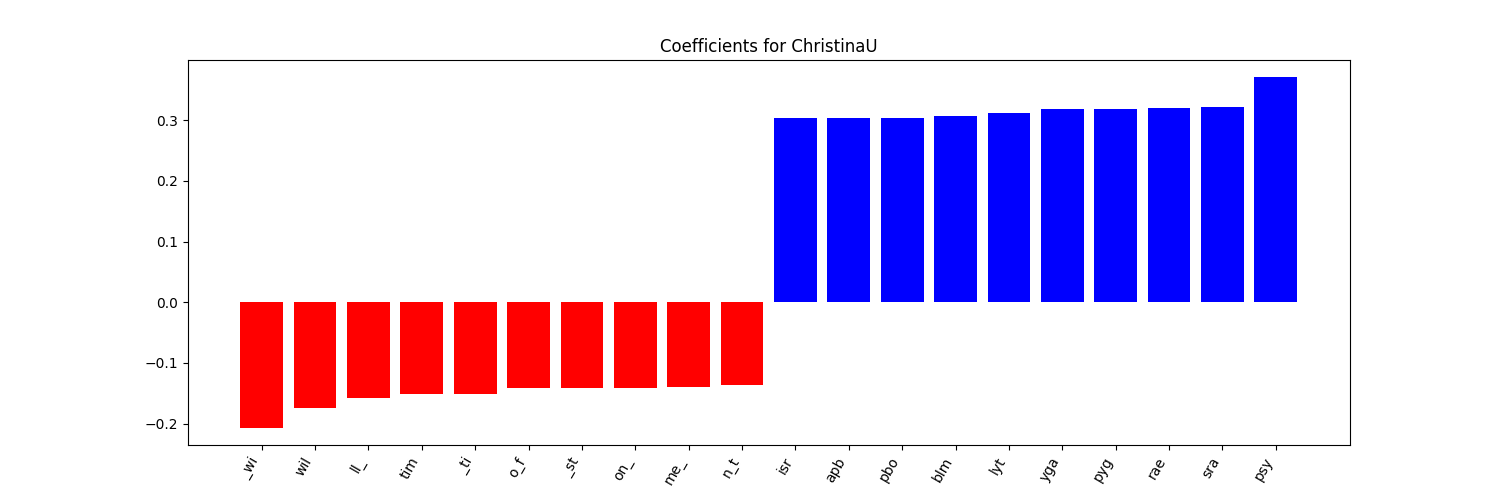}
    \includegraphics[width=0.48\textwidth]{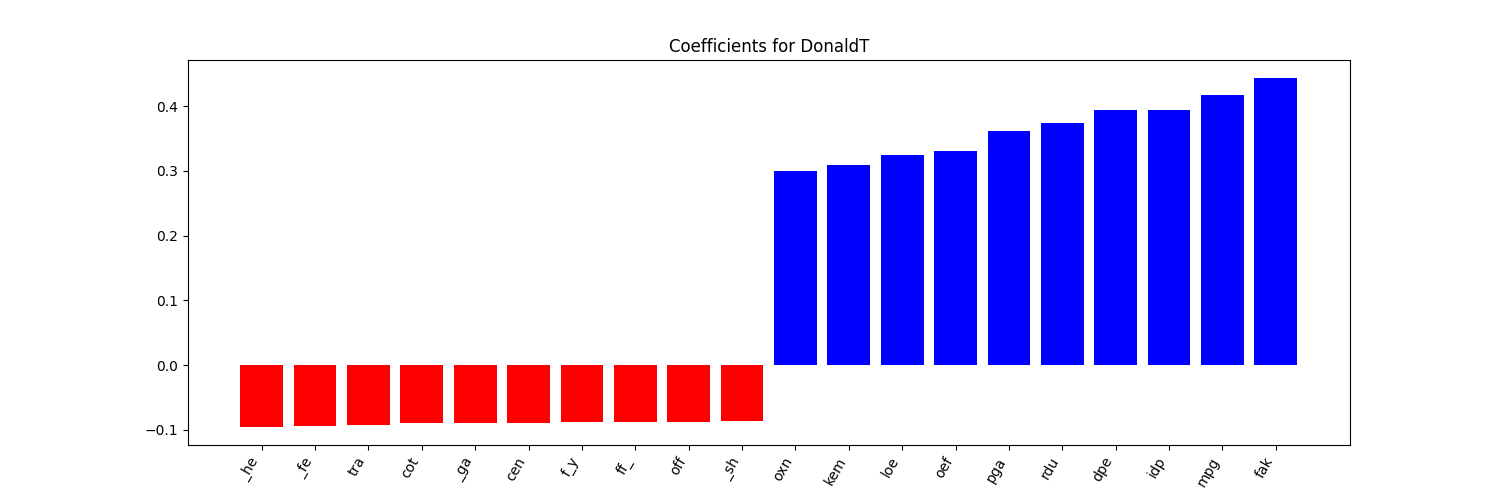}
    \includegraphics[width=0.48\textwidth]{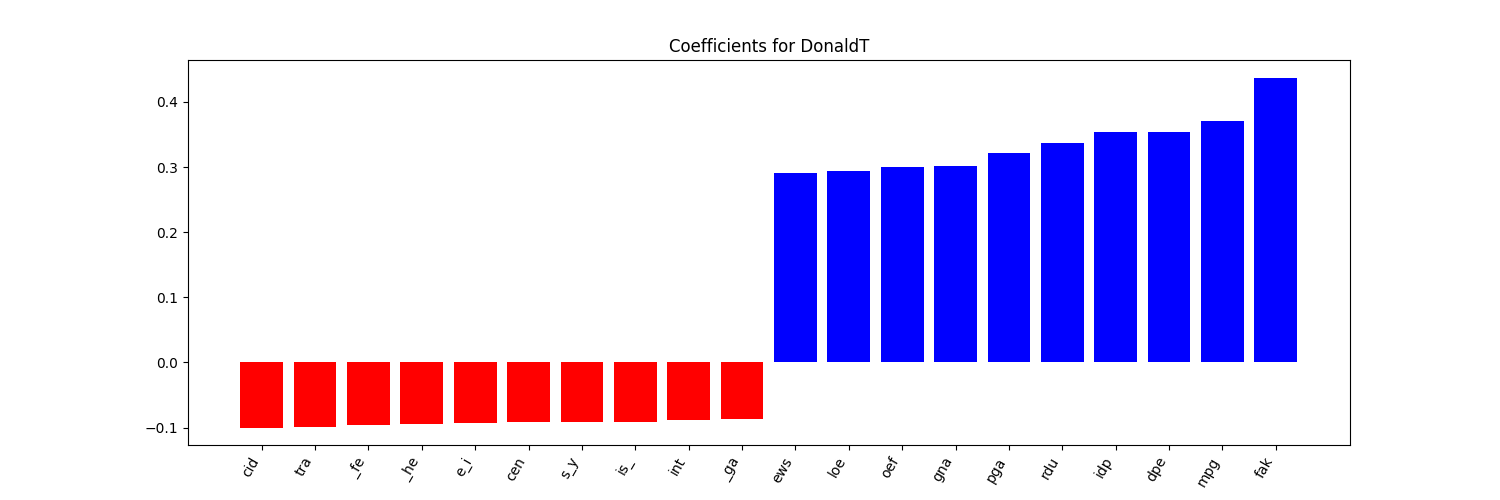}
    \includegraphics[width=0.48\textwidth]{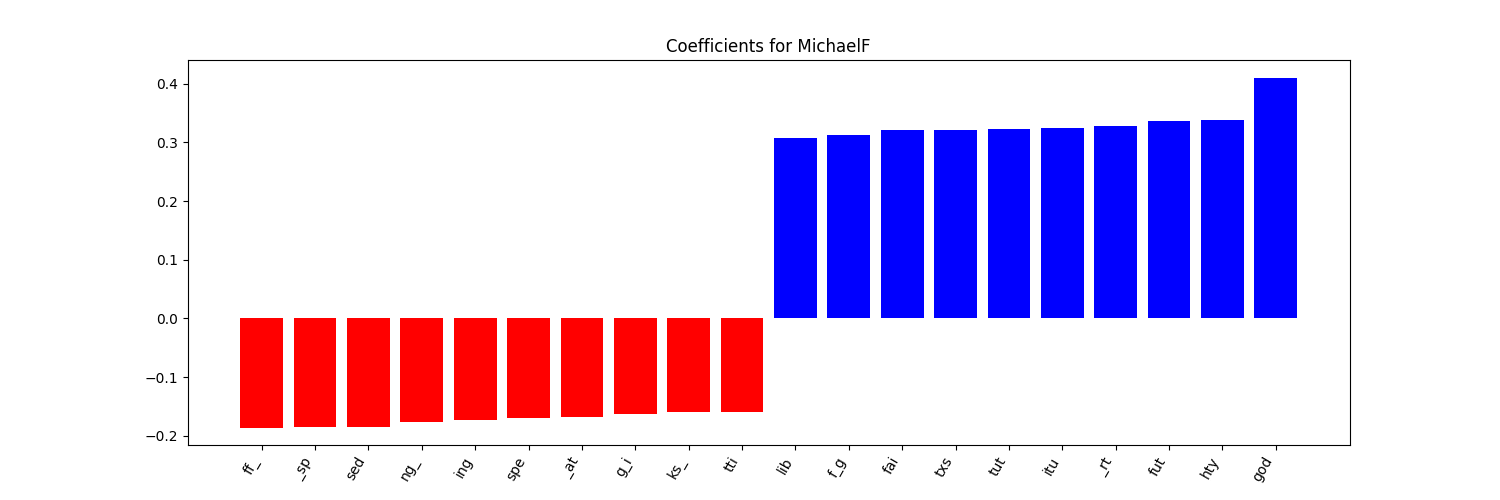}
    \includegraphics[width=0.48\textwidth]{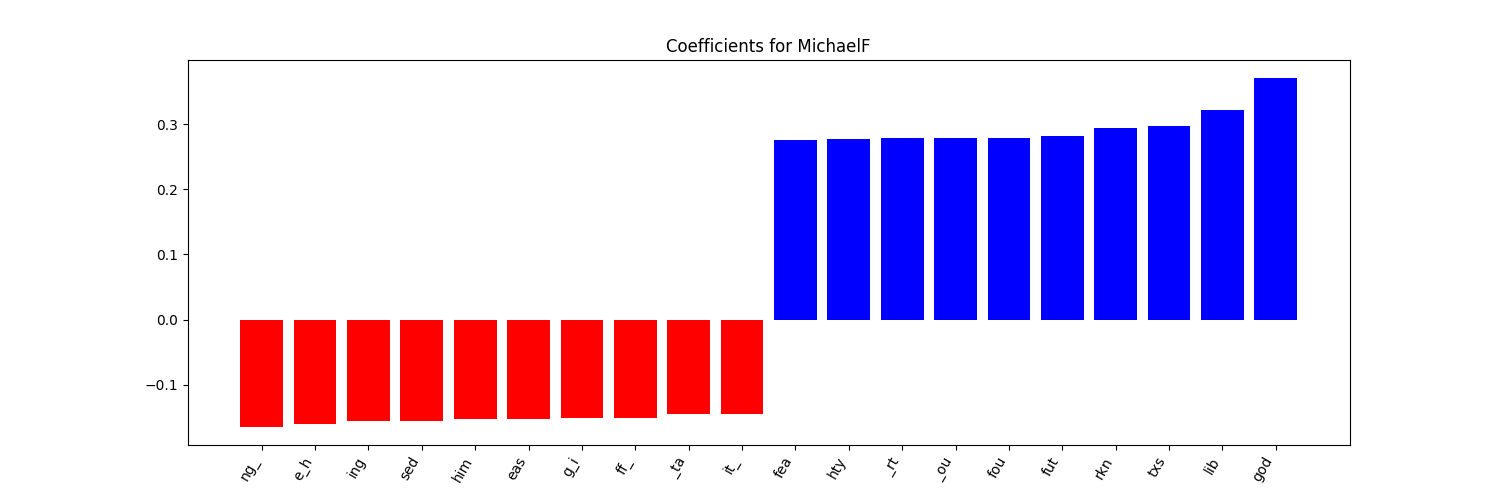}
    \includegraphics[width=0.48\textwidth]{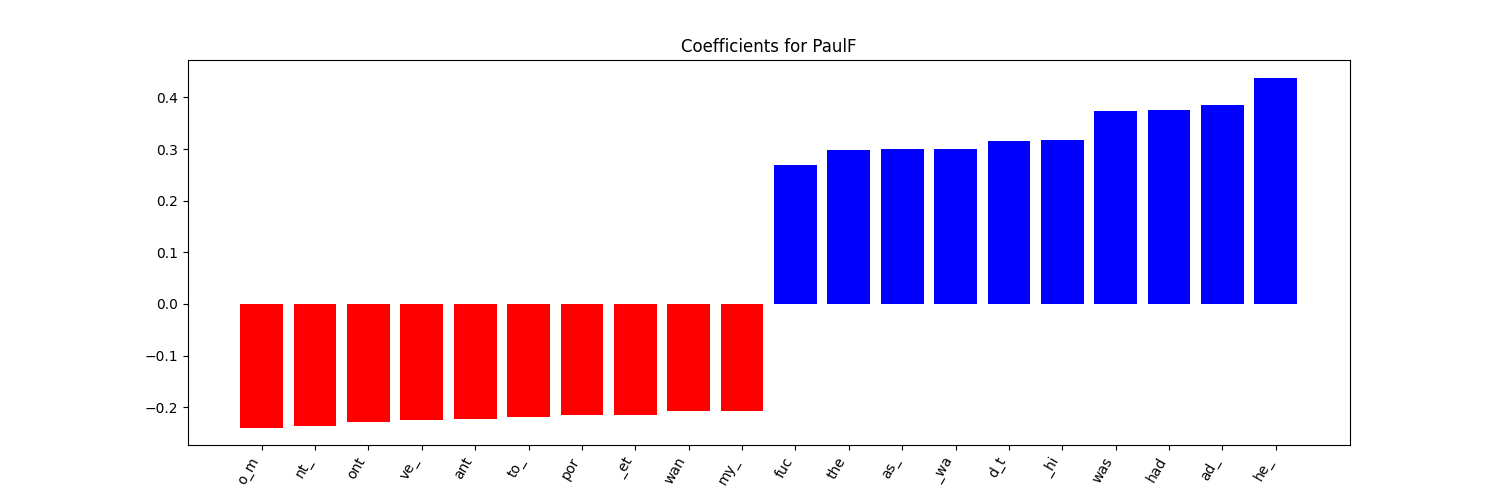}
    \includegraphics[width=0.48\textwidth]{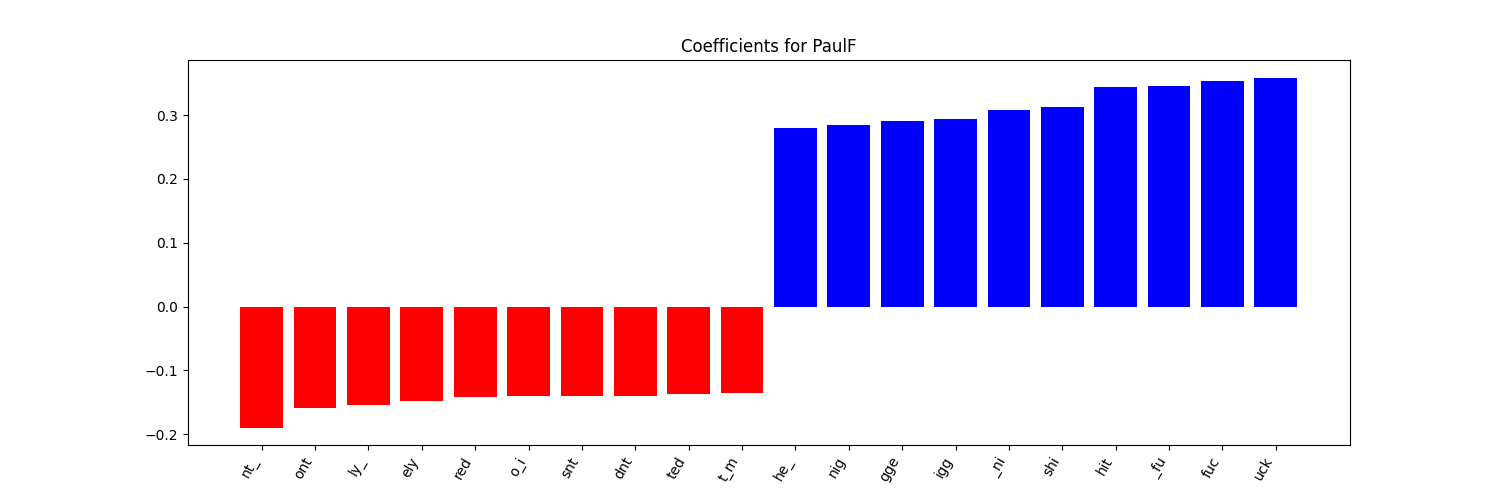}
    \includegraphics[width=0.48\textwidth]{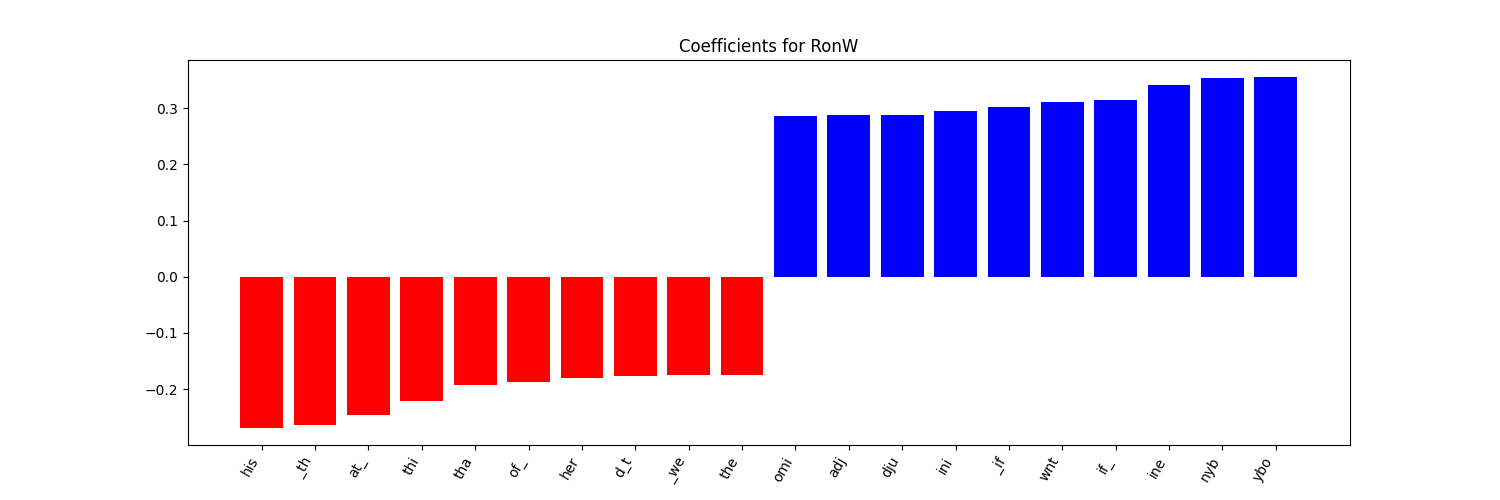}
    \includegraphics[width=0.48\textwidth]{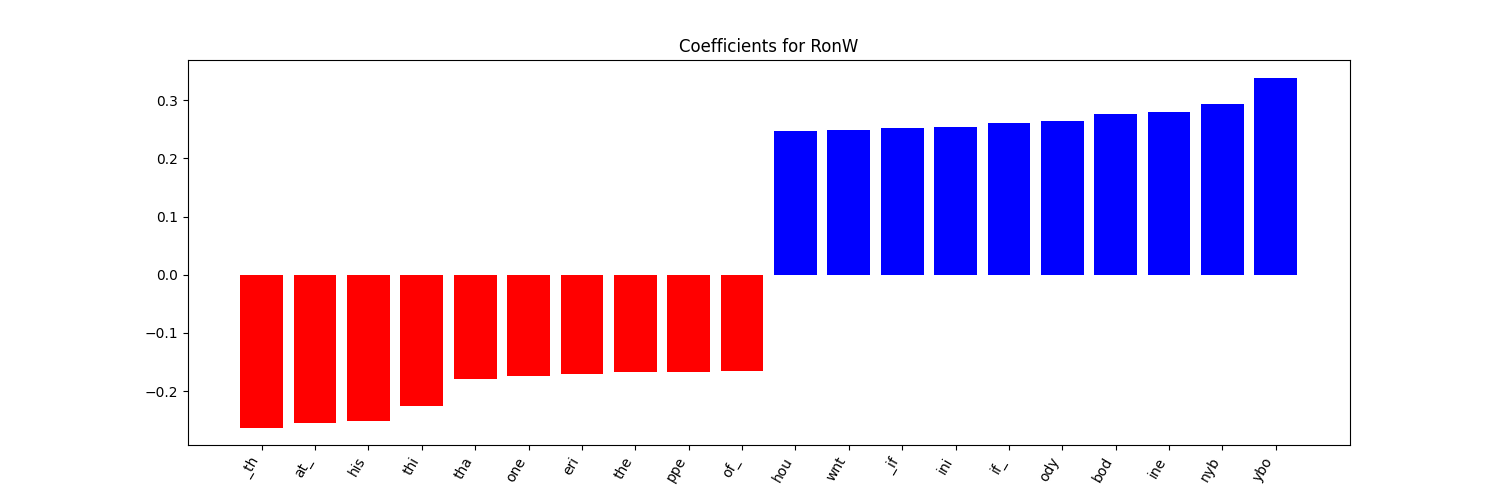}
    \caption{10 largest coefficient (negative and positive) for the Liner SVC classifiers of Christina U., Donald T., Michael F., Paul F. and Ron W., trained on the larger (left) and smaller (right) corpora}
    \label{fig:coefs}
\end{figure*}

Results obtained on the two rolling analyses, and their eventual difference, have to be contextualised by investigating the features who received the strongest coefficients in the different SVM classifiers (fig.~\ref{fig:coefs}). For some candidates, like Ron W., the features seem mostly idiolectal, like the 3-grams `nyb', `ybo' (in `anybody') or the relative avoidance of `\_th' and `his' and remain stable in between both analyses. This is also the case, for instance, for Donald T. whose most distinctive feature is `fak', part of his very idiolectal `FAKE', while other are more content related (`mpg' is even due to the regularity with which he mentioned `BrianKempGA' in the training material), a consequence of the choice of characters 3-grams as features.

For authors like Christina U., the features are very content and news-related, like the 3-grams extracted from `Israel(i)', `blm', `psy' (psychologists, psychiatrists, …), etc.

In the case of Michael F., the features seem very dependent on the small quantity of the available training material, and the grandiloquent and religious nature of the few material available, with features such as `god' (`God'), `hty' (`almighty'), `lib' (`liberty').

Finally and more importantly, these features, in their variation between analyses, give very good insight in the different results concerning Paul F. In the small corpus, due to the exclusion of his book, the most distinctive features for him are all cursory words and racist insults (`\_fu', `fuc', `uck', `shi', `hit', `\_ni', `nig', `igg', `gge', etc.); on the larger corpus, on the other hand, with the book included, they seem revealing of more neutral idiolectal (and grammatical) features, with pronouns, auxiliaries, determiners ( `he\_', `had', `was', `the', etc.).  These elements point to the larger corpus analysis being more reliable in what concerns Paul F. (especially in a cross-genre setup) than the smaller corpus analysis.

\section*{Discussion}

If the author of the Qdrops is among our candidates, the results here seem to demonstrate the major role of Ron W. in the writing of the Qdrops, at least since the switch to 8chan. The peak of Paul F. in one of the two analyses, for the period before 8chan, could very well be revealing of a real participation, even a leading role for all the period before 8chan, with perhaps afterwards a brief period of collaboration (or competition), in between the migration to 8chan and what Paul F. himself describes as the last authentic Qdrop.

Localised peaks of Christina U. or Michael F. on the other hand, while they might very hypothetically be revealing of more  occasional collaborations, should probably not be over-interpreted. Given the nature of the coefficients used by the model for them (fig.~\ref{fig:coefs}), they seem more likely to be caused by `topic similarities' due to the news and topics dealt with in the Qdrops. This could result in the choice of a similar lexicon, and even in quotations or paraphrases. Confusions between Michael F. or Christina U. on one hand, and other candidates (in training) or Q, seem due to attractions in terms of language register and generic peculiarities: samples that use a more elevated and grandiloquent type of patriotic address are brought somewhat closer to Michael F. samples, while those including heavier news-related content might be drawn towards Christina U.

\begin{figure*}[!htb]
    \centering
    \includegraphics[width=8.5cm]{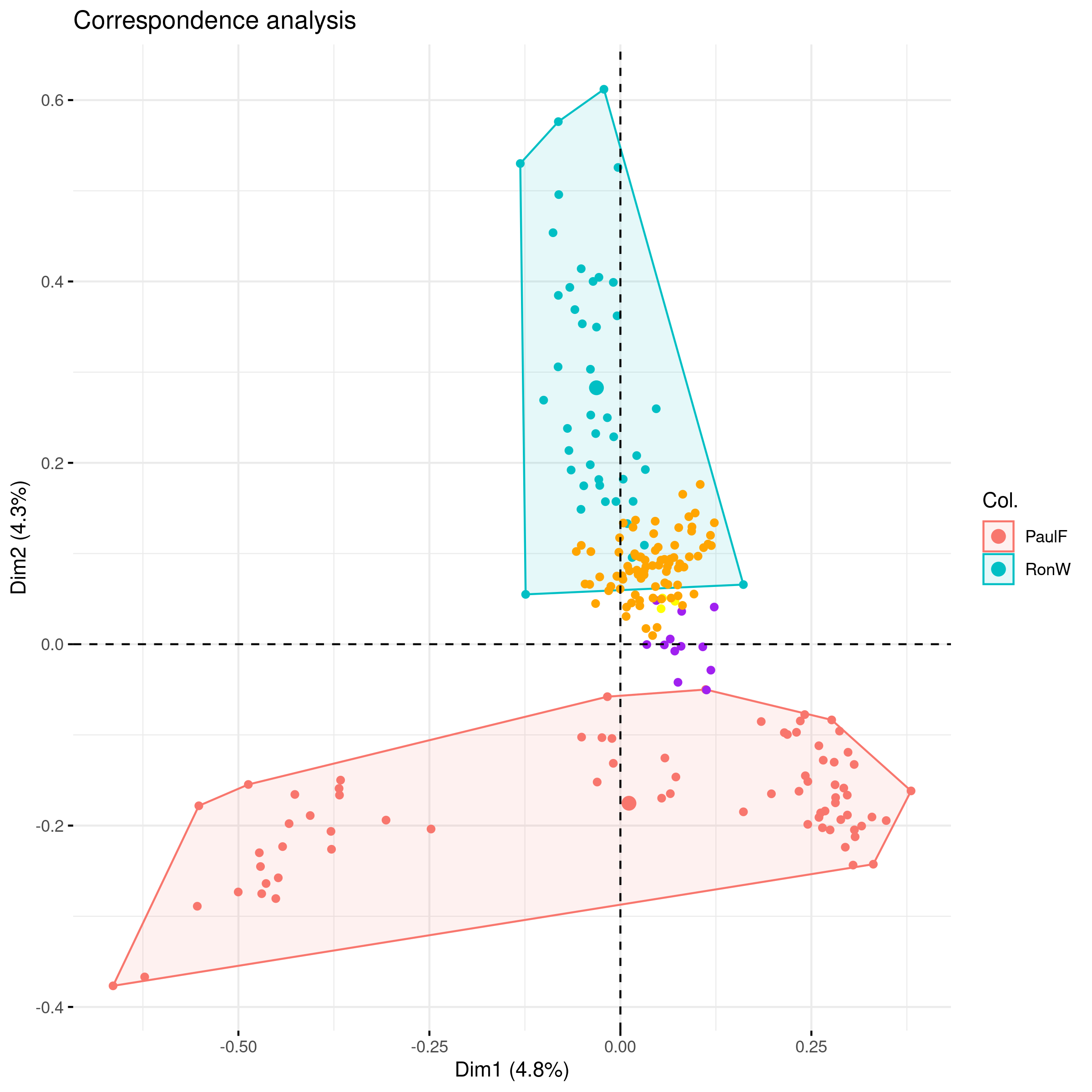}
    \caption{Dimensions 1 and 2 from a correspondence analysis of the 1000 words samples from RonW and PaulF texts; the Qdrops from the 4Chan period (purple), 8Chan period before (yellow) and after (orange) the `board compromised' post have been projected as supplementary individuals.}
    \label{fig:CA}
\end{figure*}

A more exploratory visualization of the feature space, based on correspondence analysis, shows a strong opposition (first dimension) between PaulF private posts and public writings, while the second one oppose PaulF and RonW samples (fig.~\ref{fig:CA}). Projected in this feature space, the Qdrops from the 8Chan period, after the `board compromised' post, mostly appear inside RonW data cloud, while previous posts, especially from the time of 4Chan are situated substantially closer to PaulF writings.

This paper of course has limitations. The very nature of the Q drops, a genre in itself, and the brevity of these texts, make difficult to render a finer grain picture than the one we present here. It is thus plausible, even if we cannot demonstrate it, that other punctual interventions could have occurred. The training data we collected is important, and sufficient to obtain excellent performances. Yet, more training data could of course help increase the precision and reliability of the analyses. A media outlet for instance collected the Facebook posts written by one of our candidate author, for which more support could have been helpful \citep{zadrozny_who_2018}. More generally, other individuals not listed here could of course have participated in the writing.


\subsection*{Corpus constitution}
\label{sec:corpusConstitution}
Because of their contested contents, many accounts have been deleted by
the social media platforms. Other accounts were simply deleted by the
user themselves. This raised many challenges for the corpus
constitution, which had to rely on data collection realized before the
deletion, or on difficulty searchable web archives. 
Moreover, all sources do not have the same time-span, and were created at different dates, discussing potentially different news, a source of heterogeneity in training material that creates challenge in the attribution procedure. 

We list here the sources we used for each candidate:

\paragraph{Roger S}
We collected Roger S.'s posts on Gab
(\href{https://gab.com/RogerJStoneJr}{https://gab.com/RogerJStoneJr}),
from June 14th to August 2nd, 2021.

\paragraph{Michael F}

\emph{}Michael F. wrote a series of 10 articles for the \emph{Western
Journal}, from June 29, 2020 to July 31, 2021.
\href{https://www.westernjournal.com/author/mflynn/}{(https://www.westernjournal.com/author/mflynn/}).
He also wrote a letter to ask for support to Roger S.'s wife, published
on FrankReport
(\href{https://frankreport.com/2021/06/11/banned-by-twitter-gen-michael-flynn-is-published-on-frank-report-concerning-roger-stone/}{https://frankreport.com/2021/06\-/11/ban\-ned-by-twitter-gen-michael-flynn-is-published-on-frank-report-concerning-roger-stone/}).

His article for Fox News about ISIS has also been analyzed. (\href{https://web.archive.org/web/20161215042531/http://www.foxnews.com/opinion/2016/11/02/gen-michael-flynn-after-mosul-is-liberated-isis-could-attack-us-next.html}{https://web.archive.org/web\-/20161215042531/http://www.fox\-news.com/opinion/2016/11/02/gen-michael-flynn-after-mosul-is-liberated-isis-could-attack-us-next.html}).

Finally, we collected posts on his Twitter account from October 19th, 2016 to September 18th, 2017.

\paragraph{Paul F}
We collected Paul F.'s personal writing on his website (\href{https://paulfurber.net/
}{https://paulfurber.net/}).

We retrieved archives from his twitter on account of a threadreader (\href{https://threadreaderapp.com/thread/1158540523008905216.html}{https://threadreaderapp.\-com/thread/115854052\-3008905216.html}). 

We also captured archives from the CBTS boards on 8chan, where he wrote as ``The Board Owner''.

A few posts he left on Discord were transcribed from pictures found online.

Finally, Paul F. wrote a book: \textit{Q: Inside The Greatest Intelligence
Drop In History}, included in our large corpus.

\paragraph{Jim W}
Archives of Jim W.'s Twitter account were found on archive.today
(\href{https://archive.is/https://twitter.com/xerxeswatkins}{https://archive.is\-/https://twitter.com/xerxes\-watkins}),
which preserves 9 screenshots (22 Dec 2014, 6 Feb 2016, 15 Mar 2016, 30
Mar 2016, 4 Apr 2016, 5 May 2016, 28 May 2016, 25 Mar 2017, 8 Apr 2017)
taken prior to the account suspension. A large amount of these tweets
were only citing article titles from his own media \emph{The Goldwater}.
As these titles were not necessarily (and probably not) written by him,
we chose to exclude them.

A small text was written by Jim W. on 5ch about a service problem
on 8ch that he blames on a government attack.
(\href{https://fox.5ch.net/test/read.cgi/poverty/1418027836/826}{https://fox.5ch.net/test/read.cgi/poverty/1418\-027\-836/826})

Finally, posts from his Parler account were added to the corpus.

\paragraph{Ron W}

A sample of 3130 tweets by Ron W. have been collected through the Twitter API.

We also collected his posts on Telegram from November 30th, 2021 to December, 20th, 2021.

\paragraph{Coleman R}
Under the pseudonym \emph{PamphletAnon}, Coleman R. wrote a vast
number of posts on 8chan, especially on the board ``the Storm''
(\href{https://8ch.net/thestorm/catalog.html}{https://8ch.net/thestorm/catalog.html})
of which he was the owner. We collected his posts on the Wayback
Machine, which seems to provide a complete archive of the board.

A few messages he posted on Discord (Q Central, 2017) have been archived
by DDOS
(\href{https://ddosecrets.com/wiki/Distributed_Denial_of_Secrets}{https://ddosecrets.com/wiki/Dis\-tri\-buted\_Denial\_of\_Secrets}),
at this address
\href{https://whispers.ddosecrets.com/discord/user/376607495470448643}{https://whis\-pers\-.ddo\-secrets.com/discord/user/376607495470448643}.

347 messages by PamphletAnon on Discord (Patriots' Soapbox, 2018) are
also available at:
\href{https://discordleaks.unicornriot.ninja/discord/user/41936}{https://discord\-leaks.unicorn\-riot.nin\-ja/dis\-cord/user/41936}.

We also found a small text on Reddit, where Coleman R. announces a future talk on Infowars with Rob Dew.

On September 11, 2020, Pamphlet Anon wrote a text on ``Patriots'
Soapbox'', the media he curates with Christina U. 
(\href{https://patriotssoapbox.com/opinion/memories-of-9-11-surreal-and-terrifying/}{https://patriotssoapbox.com/opinion/memories-of-9-11-surreal-and-terrifying/}).

\paragraph{Courtney T}
Collaborating with Ron and Jim W., Courtney T. publicly announced that
she knew the truth about the Q drops, and that it would be highly
disappointing to their public \citep{QHBO}. We collected her posts on Twitter, under her account IWillRedPillYou, archived here (\href{http://web.archive.org/web/20180113162029if_/https://twitter.com/IWillRedPillYou}{http://web.archive.org/web/20180113162029if\_/\-https://twit\-ter.com/I\-Will\-Red\-Pill\-You}).

\paragraph{Tracy D}
Under the pseudonym Tracy Beanz, Tracy D. published a large number of
tweets, on an account now suspended, but of which archives.today retains
92 captures
(\href{https://archive.vn/lDRyR}{https://archive.vn\-/lDRyR}),
from December 2016 to January 2021.

She also published a long post explaining herself about quarrels over Q
related publications on Steemit
(\href{https://steemit.com/drama/@tracybeanz/she-stood-in-the-storm}{https://stee\-mit\-.com/\-drama/@tracybeanz/she-stood-in-the-storm})

\paragraph{Christina U}
Christina U. wrote a number of articles on PatriotsSoapbox
(\href{https://patriotssoapbox.com/}{https://patriots\-soapbox.com/}),
of which the 5 most recent on July 7h, 2021 were collected.

We also found an online conversation she had on Muckrock  with Homeland Security on the 7th and 22nd of July 2021. (\href{https://www.muckrock.com/foi/united-states-of-america-10/patriots-soapbox-department-of-homeland-security-115196/\#file-956999}{https://www.muckrock.com/foi/united-states-of-america-10/patriots-soapbox-department-of-homeland-security-115196/\#file-956999})

Archives of her GAB profile were also analyzed.

\paragraph{Donald T}

Former president of the United States tweets were collected during the month of december 2020. We removed tweets suspected to have been written by his staff by the site \href{https://factba.se/}{https://factba.se/}.

\paragraph{Melania T}

We collected a sample of tweets by former FLOTUS thanks to the Twitter API, from January 21st, 2017 to January 19th, 2021. 

\paragraph{Eric T}

We collected a sample of 3000 tweets by the son of former POTUS thanks to the Twitter API from September 10th 2016, to June 24th 2021.

\paragraph{Dan S}

We collected a sample of tweets by the former POTUS' deputy chief of staff thanks to the Twitter API, from October 25th, 2018 to January 20th, 2021.

\subsection*{Dealing with quotations and copy/paste}

Quotes of authors outside of the corpus have been excluded as much as possible by close reading: in particular, quotes from Q, \textit{Wikipedia}, \textit{the Stanford Encyclopedia of Philosophy}, Abraham Lincoln, the Intelligence Resource Program (irp-fas), Steve Scully's biography etc. All these quotes have been removed.

Direct quotations (with or without quotation marks) and copy/paste between the writings of the different candidates can also occur. A good deal of them quote Donald, Eric or Melania T. -- Q does it too. There is also a certain number of quotations from Q by the others (such as Paul F. for instance). This could lead to small biases in the constitution of idiolectal profiles. To avoid this, we then proceeded to systematically detect citation between the candidates themselves. Direct pairwise comparison being computationally too costly for a corpus of this size, we used a Locality-Sensitive Hashing (LSH) algorithm. To that end, we used the open source \texttt{TextReuse} package \citep{textreuse}. The corpus was tokenised into sentences, and broken word bi-grams (with skip of 1, that is, allowing for any one word to be inserted between the two words of the bigrams) were counted.
For all pairs of sentences, a Jaccard similarity score was computed. Be $A$ and $B$ two samples considered as sets of bi-grams, the Jaccard similarity is computed as:
\[
J(A,B) = \frac{A \cap B}{A \cup B}
\]

All pairs of sentences with a Jaccard similarity score superior or equal to 0.5 (i.e., at least half of their bi-grams in common) were examined by a human expert, and quotations removed.

Even for $J=1$, we were sometimes confronted to false positives. Dan S. and Melania T. both use once the sentence ``we are all in this together'', without directly citing each other. We thus left this passage in both their texts. Rarely used, the sentence ``the American people are not stupid'' nevertheless appears in different texts. It was kept in the texts studied, as other simple sentences (``thank you for your service'' etc.)

Other situations were trickier to address.
For instance, Dan S. uses once the sentence: ``the best is yet to come''.  It is used five times by Q, himself quoting former President Donald Trump. This sentence could be used by anyone without directly quoting Q or Donald Trump. Yet, as its use by Dan S. starts with ``As the President says…'', we considered it a direct quotation and proceeded to deletion from Dan S.'s text.
Yet, we did not delete it from Q's own writing, as it is never used as an explicit quotation: the sentence could be used in another context, the person(s) writing the Qdrops with this sentence could try to impersonate Donald Trump, etc. In any of these cases, it would be legitimate to leave the information. Same thing goes for expression such as ``the world is watching'' or ``make America great again'', used by Donald Trump. but also by Q and some of the potential candidates here.

\subsection*{Definition of two subcorpus: dealing with generic difference and an imbalanced dataset}

The difficulties in data collection for a variety of individual and profiles, as well as the ubiquity of deleted content, not recoverable to us, forces us to adopt a a dual approach, and to build two corpora:

\begin{description}
    \item[large corpus] in which we include the larger number of candidates, whatever the number of samples available to us and the genre of said samples. It contains everything described in the \textit{contains everything described in the Corpus constitution} section above;
    \item[controlled corpus] in which we removed authors for which only too cross-genre and/or too few samples are available, and do not include training material that is too different from the rest (in particular, books). It is the same as the previous one, minus 
    \begin{itemize}
        \item interviews transcripts (Michael F., Paul F.);
        \item a book by Paul F.;
        \item the small amount of available data for Courtney T. and Roger S.
    \end{itemize}
\end{description}

In both cases, due to the limitations in data collection and available material, the quantity of training material is imbalanced between authors, a potential problem in machine learning. To counter this effect, we used class weights during training,
where errors for a given class are penalised not always by one, but by a specific weight inversely proportional to class size, where the weight for class $i$ is computed as
\[
w_i = \frac{N}{C (n_i + 1)}
\]
where $N$ is the total number of samples, and $C$ the total number of unique classes
and $n_i$ the number of samples in class $i$.
 Xe used the \texttt{sklearn} `balance' implementation \citep{scikit-learn}.

\subsection*{The genre of ``Q drops'': a methodological challenge}

The study of the Q drops raises a number of specific challenges. First
of all, the Q drops constitute per se a kind of a \emph{genre}. It
follows specific rules that most people would not use in another
context: they do not look like a regular blog, media or social media
post, nor do they belong to any specific literary genre etc. This forces
us to consider our attribution problem as a cross-topic attribution
problem.

This specific genre has consequences on many linguistic properties of
the Q drops. The structural brevity of the sentences for instance
prevents us from taking the sentence lengths as a clue of who
wrote what. The overwhelming proportion of interrogative forms,
especially in the first Q drops, makes it difficult to reason on
morpho-syntactic sequences, as they are often extremely and
artificially stereotyped.Part-of-Speech n-grams such as  ``Interrogative
pronoun - conjugated verb - common noun'' would for instance emerge as a
signature of the Q drops, and could derail our analysis, by pointing to
any of the suspects using interrogative forms the most in other
contexts. Finally, the elliptic style of the Q drops, written almost as
if they were a telegram, distorts the use of function words, less
frequent than expected in the Q drops corpus. Approaches relying only on function words could be made less robust by this distortion.

We thus chose to work on character trigrams, the most flexible and
reliable feature we could use in the very specific context of this
study, and a widely acknowledge feature in stylometry, in particulary for its supposed capacity to capture grammatical morphemes \citep{kestemont2014function,sapkota2015not}
while bearing in mind potential greater sensibility to thematic attractions in comparison to function words.

\subsection*{Detecting style changes: rolling stylometry}

Collaborative writing is not necessarily easy to handle. The scenario in which authors simply took turns, and divided the work between themselves is already complicated to address. But when the collaboration is more complex, especially when the various authors contribute together to the same passages, the style of the original authors can be hard to recognize. The collaboration then results in a new style, that does not match the style of one of the authors \citep{kestemont2015collaborative}.

The principle of rolling stylometry \citep{eder2016rolling} is simple: rather than attributing a whole text, we arbitrarily decompose it in a series of overlapping smaller parts: from the 1st word to the 1001th word, from the 2nd word to the 1002nd word etc. Then, we attribute each of these parts to a certain author. We only have to define the length of these parts, and by how much they overlap.

Rolling stylometry has been successfully implemented in a wide variety of settings. With Burrows' delta, it has for instance been used to assess Ford's claims about his implications in collaborations worth Joseph Conrad \citep{rybicki_collaborative_2014}, to determine the beginning of Vostaert's intervention on Dutch Arthurian novel \emph{Roman van Walewein} \citep{van2007delta}, or to understand Lovecraft's and Eddy's implication in \emph{The Loved Dead} \citep{gladwin2017stylometry}. Using support-vector machines, rolling stylometry more recently helped to confirm Fletcher and Shakespeare's collaboration for \emph{Henry VIII} \citep{plechavc2019relative} or Molière and Corneille's collaboration for \textit{Psyché} \citep{cafiero2021psyche}.

\subsubsection*{Support Vector Machine}

We choose to train linear Support Vector Classifiers (SVC) to identify the style of each potential candidate. The family of Support Vector Machines algorithm has been widely and successfully used for authorship attribution in a variety of settings and languages, and for very diverse sources, ranging from e-mails or blogs to Shakespeare plays \citep{de2001multi, diederich2003authorship, mikros2012authorship, ouamour2012authorship, marukatat2014authorship, plechavc2019relative}. At the PAN competition, a reference for digital text forensics and stylometry, it also served as a baseline for the ``cross-domain authorship attribution'' tasks the last time they were proposed in 2018 \citep{stamatatos2018overview} and 2019 \citep{kestemont2019overview}. The Q drops being a sort of ``domain'' in their own, our own task can be considered as a cross-domain authorship attribution task. Other classifiers could have been used, but would not offer the interpretability we need for that kind of task. In such a delicate context, being able to get a simple and clear intuition on which features the classifier relies is simply crucial. If the features selected were to be too much related to the content of the texts, rather than to properly linguistic properties of the person's discourse,  we need to be aware of it, and train more properly a new classifier.

To determine the choice of features and the size of the training samples, we are constrained by two antagonistic goals: the shorter the samples, the more detailed and precise the results that we will get in terms of attribution, yet the longer, the more statistically reliable. Particularly, authorship attribution has proven to require relatively high amounts of data, with a floor for reliable authorship attribution between 1000 and 3000 words,
depending on genre and language \citep{eder_does_2015,eder_short_2017}. 
The question of sample length can also be linked to the
difficulty of the attribution task; cross-domain attribution with
multiple candidates presents a challenge in this regard.

On the other hand, the features we retained, character 3-grams, could increase robustness, ad they are known to reduce sparsity and perform well in attribution studies  \citep{kestemont2014function,sapkota2015not}.  While punctuation can strongly reflect authorial signature \citep{sapkota2015not}, we removed led to remove it 
because of variety of platforms from where the data were recovered could cause inconsistencies in the use of signs that can be encoded in different fashions, e.g., apostrophes.

For these reasons, we retain a setup that is a compromise between reliability and finer grain analysis:
\begin{description}
    \item[Sample length] 1000 words;
\item[features] character 3-grams (all, including punctuation).
\end{description}

To evaluate our setups, we opt for a leave-one-out cross evaluation on the training corpus (Table~\ref{tab:cross-evaluation}), for a combination of reasons. First, some of the samples are of relatively small size in the context of training an efficient classifier. A leave-one-out procedure helps us using the maximal amount of data. The constitution of which relevant and coherent training sets is also a question in our case. In that context, leave-one-out evaluation can provide a more robust estimate of the model's performance, as it accounts for the potential variability caused by the specific sample chosen for testing. Finally, leave-one-out can help us avoid overfitting \citep{ng1997preventing, ghojogh2019theory}, which is a danger in our study - we do not want the classification to be based on some specific piece of news of particular interest to one of the candidates, but to have reliable classification based on linguistic features appearing in a large range of contexts.

The confusion matrix gives more information on the nature of the small number of classification errors (Table~\ref{tab:confMat}). 
As can be expected, performance is slightly lower for authors for which training material is very limited (ColemanR, CourtneyT, RogerS). For the others, it is above 95\%, if we except a few confusions between Michael F. and Roger S. (on the large corpus only). These can be explained by the limited size of Roger S. training data, thematic attractions, the fact that he talks about Flynn more or less directly, but also probably by a generational (age) bias.

\begin{table*}[ht]
   \centering
\caption{Results of the leave-one-out cross-evaluation for the large corpus (left) and the controlled corpus (right).}
\tiny%
    \begin{tabular}[t]{c|rrrr}
&\textbf{precision}&\textbf{recall}&\textbf{f1-score}&\textbf{support}\\ \hline \hline
ChristinaU&1.00&0.93&0.97&15\\ 
ColemanR&0.89&1.00&0.94&8\\ 
CourtneyT&1.00&0.67&0.80&6\\ 
DanS&1.00&1.00&1.00&9\\ 
DonaldT&1.00&1.00&1.00&9\\ 
EricT&1.00&1.00&1.00&26\\  
JimW&1.00&1.00&1.00&24\\ 
MelaniaT&1.00&1.00&1.00&25\\ 
MichaelF&1.00&0.84&0.91&19\\ 
PaulF&0.96&1.00&0.98&74\\ 
RogerS&0.75&1.00&0.86&6\\ 
RonW&0.98&1.00&0.99&43\\ 
TracyD&1.00&0.94&0.97&18\\ \hline
\textit{accuracy}&&&0.98&282\\ 
\textit{macro avg}&0.97&0.95&0.96&282\\ 
\textit{weighted avg}&0.98&0.98&0.97&282
    \end{tabular}
    \begin{tabular}[t]{c|rrrr}
&\textbf{precision}&\textbf{recall}&\textbf{f1-score}&\textbf{support}\\ \hline \hline
ChristinaU&1.00&1.00&1.00&15\\ 
ColemanR&1.00&0.88&0.93&8\\ 
&&&&\\ 
DanS&1.00&1.00&1.00&9\\
DonaldT&1.00&1.00&1.00&9\\ 
EricT&1.00&1.00&1.00&26\\ 
JimW.&1.00&0.92&0.96&12\\ 
MelaniaT&1.00&1.00&1.00&25\\
MichaelF&1.00&0.94&0.97&17\\ 
PaulF&1.00&0.96&0.98&24\\ 
&&&&\\ 
RonW.&0.93&1.00&0.97&43\\ 
TracyD&0.95&1.00&0.97&18\\  \hline
\textit{accuracy}&&&0.98&206\\ 
\textit{macro avg}&0.99&0.97&0.98&206\\ 
\textit{weighted avg}&0.98&0.98&0.98&206
    \end{tabular}
   \label{tab:cross-evaluation}
\end{table*}

\begin{table*}[htbp]
    \centering
\caption{Confusion matrix for the leave-one-out evaluation on the larger corpus}
\tiny %
\begin{tabular}{c|rrrrrrrrrrrrr}
\textit{Predicted}&ChristinaU&ColemanR&CourtneyT&DanS&DonaldT&EricT&JimW&MelaniaT&MichaelF&PaulF&RogerS&RonW&TracyD\\ 
\textbf{Expected}&&&&&&&&&&&&&\\
ChristinaU&14&0&0&0&0&0&0&0&0&\textbf{1}&0&0&0\\ 
ColemanR&0&8&0&0&0&0&0&0&0&0&0&0&0\\ 
CourtneyT&0&0&4&0&0&0&0&0&0&\textbf{1}&0&\textbf{1}&0\\ 
DanS&0&0&0&9&0&0&0&0&0&0&0&0&0\\ 
DonaldT&0&0&0&0&9&0&0&0&0&0&0&0&0\\ 
EricT&0&0&0&0&0&26&0&0&0&0&0&0&0\\ 
JimW&0&0&0&0&0&0&24&0&0&0&0&0&0\\ 
MelaniaT&0&0&0&0&0&0&0&25&0&0&0&0&0\\ 
MichaelF&0&\textbf{1}&0&0&0&0&0&0&16&0&\textbf{2}&0&0\\ 
PaulF&0&0&0&0&0&0&0&0&0&74&0&0&0\\ 
RogerS&0&0&0&0&0&0&0&0&0&0&6&0&0\\ 
RonW&0&0&0&0&0&0&0&0&0&0&0&43&0\\ 
TracyD&0&0&0&0&0&0&0&0&0&\textbf{1}&0&0&17
    \end{tabular}
\label{tab:confMat}
\end{table*}

We then apply our models to all successive overlapping slices of Q
drops, arranged in chronological order, with window size of length 1000 words and step 200 words. We then plot the resulting decision functions for each classifier. The higher the value, the more likely
the attribution of a sample to a given author

All analyses are implemented in Python, inside the \emph{SuperStyl}
package \citep{camps_supervised_2021}, and use internally the SVM and pipeline facilities provided by \emph{Sklearn} \citep{scikit-learn}. Plots are created using R and Python (\texttt{matplotlib}). 

\subsubsection*{Correspondence analysis} Correspondence analysis \citep{Benzecri_corresp_1973} has been performed on a contingency table of the 1000 words samples by RonW and PaulF based on 4105 character 3-grams selected for statistically reliability based on a procedure previously described \citep{cafiero2019moliere}. The QDrops have then been inserted as supplementary rows, using the implementation in the R package FactoMineR \citep{le_factominer_2008}. 
The significance of the two first axes remains relatively low, in part because of the high dimensionality of the input table, but the data clouds for PaulF and RonW are clearly separated, and the QDrops appear in an intermediary position.

\section*{Ethical statement}

Even if written independently, this study tried to abide as much as possible to the principles of the ``\textit{Pratiquer une recherche intègre et responsable}'' guide by the \textit{Centre National de la
Recherche Scientifique}'s ethics board \citep{comite_dethique_du_cnrs_pratiquer_2017}. This article does not reveal the
identity of individuals that were not broadly known beforehand. In
this case, all candidate authors were already either public figures, or
individuals whose identity had been stated in major media outlets (NBC,
HBO etc.). It only uses information that was in conscience made publicly
available by the candidate authors and was accessible through
standard internet searches at the time of data collection.

For ethical reasons and to respect the privacy policy of the platforms studied here, we do not freely release any content studied here. To respect the data sharing and open data principles, we however detail our data collection method, which should be sufficient to ensure reproducibility in most cases. Some contents could not be available anymore when an attempt at reproducing our computations is performed. In that case, these missing materials could be delivered to research teams on request.

We choose to designate the candidates we study only by their first name and initial, not to impact internet searches on their names. 

Finally, we remind that this paper does not assert in any way that other persons outside of the persons studied here could not have written the Q drops.

\subsection*{Acknowledgements}

The authors would like to thank 
 David D. Kirkpatrick from the \textit{New York Times} and Frederick Brennan for their help during this investigation. Errors remain our own.
\textbf{Funding}: no funding was used for this research.
\textbf{Data and materials availability}: Code is available on Zenodo, doi: \href{http://dx.doi.org/10.5281/zenodo.6164620}{10.5281/zenodo.6164620}.
Data available on request.

\bibliographystyle{plainnat}
\bibliography{biblio}

\end{document}